# MOMO – Deep Learning-driven classification of external DICOM studies for PACS archivation

Frederic Jonske[1*], Maximilian Dederichs[2], Moon-Sung Kim[1], Jan Egger[1], Lale Umutlu[2], Michael Forsting[1], Felix Nensa[1,2], Jens Kleesiek[1]

[1] Institute for AI in Medicine (IKIM), University Medicine Essen, Germany.
[2] Institute of Diagnostic and Interventional Radiology and Neuroradiology, University Medicine Essen, Germany.
[*] Corresponding author

## Abstract

Patients regularly continue assessment or treatment in other facilities than they began them in, receiving their previous imaging studies as a CD-ROM and requiring clinical staff at the new hospital to import these studies into their local database. However, between different facilities, standards for nomenclature, contents, or even medical procedures may vary, often requiring human intervention to accurately classify the received studies in the context of the recipient hospital's standards. In this study, the authors present **MOMO** (**MO**dality **M**apping and **O**rchestration), a deep learning-based approach to automate this mapping process utilizing metadata substring matching and a neural network ensemble, which is trained to recognize the 76 most common imaging studies across seven different modalities. A retrospective study is performed to measure the accuracy that this algorithm can provide. To this end, a set of 11,934 imaging series with existing labels was retrieved from the local hospital's PACS database to train the neural networks. A set of 843 completely anonymized external studies was hand-labeled to assess the performance of our algorithm. Additionally, an ablation study was performed to measure the performance impact of the network ensemble in the algorithm, and a comparative performance test with a commercial product was conducted. In comparison to a commercial product (96.20% predictive power, 82.86% accuracy, 1.36% minor errors), a neural network ensemble alone performs the classification task with less accuracy (99.05% predictive power, 72.69% accuracy, 10.3% minor errors). However, MOMO outperforms either by a large margin in accuracy and with increased predictive power (99.29% predictive power, 92.71% accuracy, 2.63% minor errors). Here we show that deep learning combined with metadata matching is a promising and flexible approach for the automated classification of external DICOM studies for PACS archivation.

**Keywords**: Machine Learning, Artificial Intelligence, PACS, DICOM, Medical Image Classification.

# 1 Introduction

Machine Learning has had a large impact on the field of medical studies, particularly in radiology [1]. Great strides have been made in the automated assessment of medical images in decision making [2], prediction [3], or diagnostics [4]. An often overlooked application of deep learning is the elimination of repetitive tasks in the clinical routine. One of these tasks, currently requiring several dedicated medical-technical personnel, is processing and archivation of external DICOM studies. Frequently, patients from other facilities will hand in CD-ROMs containing imaging studies, which are archived in the local hospital's PACS. The origin facility of these studies occasionally has a different language, naming standard for procedures, differently composed imaging studies, or even different procedures. Adherence to DICOM standards, such as clean labeling, file structure, and ordering (see for example [5]), is not always guaranteed, and Güld et al. [6] have found that DICOM metadata is often unreliable (the DICOM tag *Body Part Examined* was found to be incorrect in 15.3% of cases). Studies that are thus incorrectly mapped to the local study nomenclature cause problems, from medical staff being unable to identify the relevant study for diagnostics and treatment, to arranging for a procedure to be performed unnecessarily. The lack of adherence to established standards additionally gives rise to a compositionality problem - a single study can contain multiple different series which together comprise one class, and such a composition can be incompatible with the recipient hospital's classification scheme.

Related works have performed body region classification using various deep learning strategies (see [7], [8], [9], [10] and [11]), but none with the explicit goal of study classification for PACS archivation in mind, nor for this comprehensive list of modalities.

Similarly to Dratsch et al. [7], who used a neural network for evaluation of radiographs in the same context, the authors of this paper propose to extend the scope of the application onto multiple modalities at once - CT, X-Ray Angiography, Radiographs, MRI, PET, Ultrasound, and Mammograms. Firstly, we aim to provide an automated classification algorithm, which can be integrated into the clinical routine, to help automatically organize the local PACS, as suggested in [7]. Secondly, we aim to establish the usefulness of neural networks in this context, as a part of or as a standalone solution. In this study, we propose a deep learning-based approach and compare its performance to a commercial product.

# 2 Materials and methods

## 2.1 Training images

11,934 de-identified imaging series were retrieved from the local PACS, covering random timeframes and patients, and 76 different types of studies, with around 150-200 imaging series for each class or fewer if unavailable. These classes (see Supplementary Materials) comprise the most common study types at our hospital. Imaging series were automatically labeled according to the class they were saved as in the PACS. Images were only rejected based on their series descriptor (excluding non-representative series such as topograms), but not quality or demographics. This is justified, as the external test set might feature low-quality scans and previously unseen image compositions. 10% (each) of the images are randomly drawn from the internal dataset to create a validation and test set for performance evaluation of the networks.

## 2.2 External studies

843 external studies (covering studies arriving at the local hospital from January 4th, 2021 to January 8th, 2021), were labeled by a radiographer of the medical-technical staff with several years of experience as a radiographer team leader. Every study received at least one label, and some studies received several labels if they could be reasonably interpreted as multiple study types. This external test set was not used during training or for fine-tuning. Accuracy and predictive power on this set yield the final performance measures.

## 2.3 Image preprocessing and neural network training

We evaluated two neural networks, a Resnet-152 [12] and DenseNet-161 [13], both pre-trained on ImageNet [14]. Images fed into the networks were first normalized into the range [0,1]. All image series were reduced to single-channel, greyscale images, and resampled to 512x512 along the X- and Y-axis, using spline interpolation. Two-dimensional image series became 512x512x3 images by stacking the first, middle, and last layer along the Z-axis. Three-dimensional images were resampled to length 512 along the Z-axis. The 512x512x3 input images were created by constructing the maximum intensity projections (MIPs) along the major axes. To avoid gross resampling errors, images where $Z<40$ were treated as two-dimensional.

Images were randomly flipped or rotated during training (with probability p=0.2), as image orientation can vary on external images. Performance was evaluated using cross-entropy loss. The networks were trained using transfer learning, freezing every layer except the final layer and the classifier (for hyperparameters, see Supplementary Materials). To eliminate unnecessary cross-contamination, we created an independent network for every modality except PET. For studies containing PET and CT or PET and MRI images, the CT/MRI images were evaluated by the respective network. The PET images themselves were not used during training or inference, as pure PET images are unsuitable for body site recognition.

## 2.4 Temperature Scaling

At inference time, the logits (the raw output of the network's final layer) are additionally scaled with a constant factor $T$, to minimize the average expected difference between the network-predicted (confidence) and true probability (accuracy). The holdout validation set was used to fit this scaling factor and we performed this procedure for every network we trained. To accomplish this, all N predictions were divided into M equally spaced confidence bins $B_m$ and $T$ was chosen to minimize the expected calibration error (ECE):

$$ECE = \sum_{m=1}^{M} \frac{|B_m|}{N} * |\, accuracy(B_m) - confidence(B_m)\,|. \qquad (1)$$

This process is called temperature scaling and guarantees that the returned confidences of the neural network are statistically meaningful probabilities. For example, a network prediction with a confidence of 0.8 should be correct approximately 80% of the time. A network that has this property is called "calibrated". This behavior is desired so the network can differentiate between confident and uncertain predictions, which can then be treated differently. This calibration increases the performance of our algorithm later on (see *Experiment 2*). For more information see Fig. 1 and the original paper on temperature scaling [15].

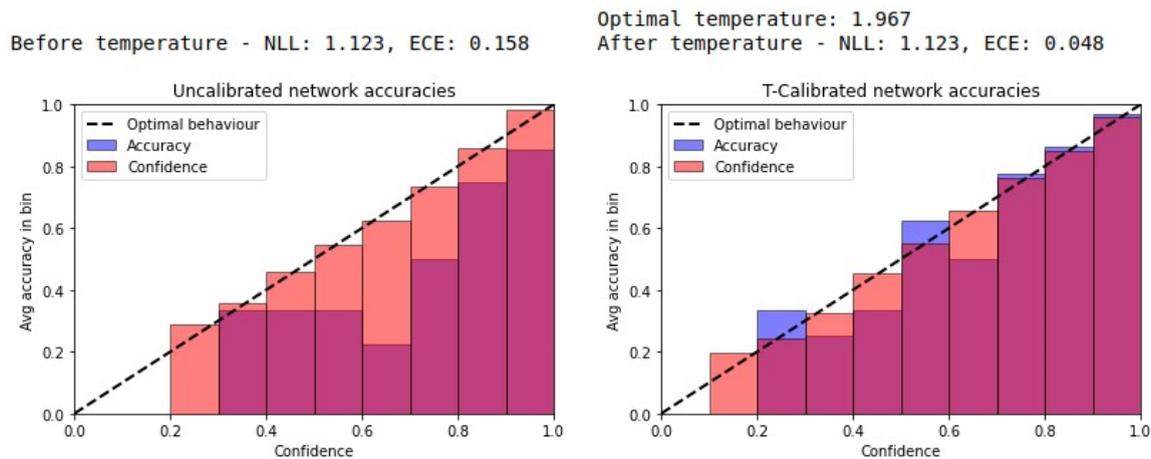

Figure 1: Shown is an example temperature scaling, performed using 10 bins equidistant in probability. The averaged confidences are transparent red while the accuracy is transparent blue. An excess of red indicates that the neural network is overconfident in its predictions. The left plot shows the confidence and accuracy before temperature scaling was applied, the right plot after. Notably, the expected calibration error (ECE) has gone down significantly, while the negative log-likelihood loss (NLL) has remained exactly the same. This is intuitive, as temperature scaling never changes the actual predictions themselves, only the network's confidence in them.

## 2.5 MOMO (MOdality Mapping Orchestration)

MOMO's algorithm follows a hierarchical structure (see Fig. 2). During each step, MOMO attempts a prediction. If successful, it returns the prediction, otherwise it proceeds to the next step. It begins by matching the study's *Procedure Code* (DICOM tag) against known Procedure Codes. Next, the *Study Description* tag is evaluated in the same manner. If both fail, MOMO attempts to partially match the *Study Description*, by finding the longest matching substring with a minimum length of 6 characters (or 4 characters for exact matches of short keywords) between it and a list of pre-specified keywords. Every such keyword is associated with a prediction, which is returned if a match is found. After that, it extracts every pre-specified metadata key from every series in the study and attempts to partially match all of them against this list. Every match is weighted equally, with a simple majority choosing the prediction. A rules system disallows or modifies some of these votes based on a configuration file (see Supplementary Materials). On a tie (or no votes), all imaging series are resampled and fed into the neural network corresponding to their modality. Each network prediction yields one vote, weighted by the prediction's confidence. All votes are summed and evaluated. The validity of combining information in this voting approach is established in a Monte Carlo experiment using pseudo data. (see *Experiment 2*).

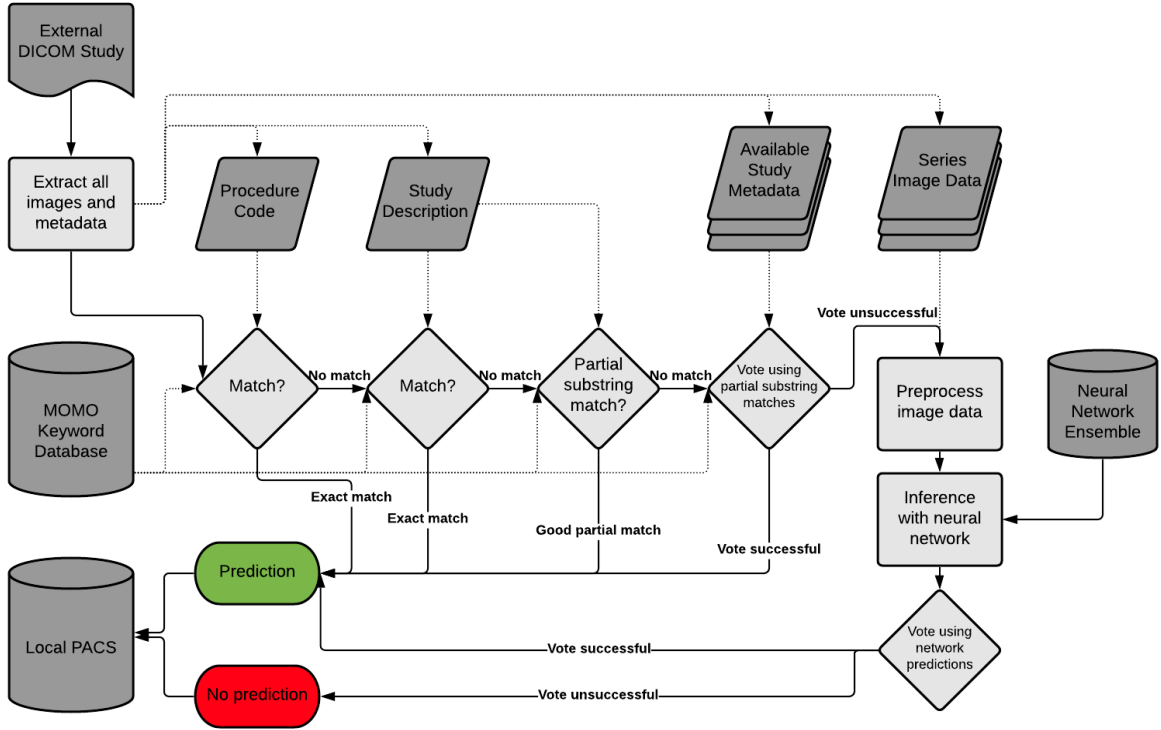

Figure 2: MOMO extracts meta- and imaging data and then progresses through its decision layers (light gray diamonds) one by one until a decision is made. The voting approach used for combining information is detailed in the Methods section. The entire algorithm can be modified from the Modality Mapping Database (a single tabular reference document containing all study classes and associated keywords), and a single configuration file (which holds technical parameters).

## 2.6 Experiments

1) To assess the performance of the neural networks on a per-series basis, their accuracies are calculated for the internal test set, using five-fold Cross-Validation. Accuracies are reported with 95% confidence intervals.

2) As a proof of concept, we include a Monte-Carlo experiment, in which a statistically ideal version of the evaluation performed by MOMO is created using pseudo data, which shows the validity of the network voting and the value of the calibration performed earlier. To simulate a calibrated network with accuracy $\alpha$, we must guarantee over every series $m \in 1, ..., M$ and study $n \in 1, ..., N$ that both:

$$\frac{\sum_{n=1}^{N} \sum_{m=1}^{M} P(x_{n,m})}{M*N} = \alpha \text{ with } P(x_{n,m}) = \{1 \text{ if prediction } x_{n,m} \text{ is true, else } 0\} \qquad (2)$$

and

$$\sum_{m=1}^{M} \frac{|B_m|}{N} * | \, accuracy(B_m) - confidence(B_m) \, | \qquad (3)$$

is small (meaning the calibration approximately holds for any arbitrary choice of bins). Therefore, we first choose a random threshold $T$, drawing from a beta-distribution, where $P$ and $Q$ satisfy $P+Q=4$ and $\alpha$ can be freely chosen. This parameter choice, while arbitrary, nicely distributes the mass of random variables in the range [0,1], while retaining a peak near the mean of the distribution. Next, we draw a random variable $P$ from a uniform distribution in [0,1]. If $P<T$, a vote for the correct pseudo label is saved. If $P>T$, one of 11 random false pseudo labels is saved instead. The vote is assigned the weight $T$. This process is repeated several times per pseudo study before a simple majority of the weighted votes decides the prediction. $N=1,000,000$ pseudo studies are simulated in this manner. The experiment is performed for both uncorrelated and maximally correlated false predictions. In the latter case, each true label has exactly one associated false label which is always saved on a false prediction. These conditions simulate an ideal environment and neglect issues such as compositionality or different domains from which studies could come, where prediction performance might systematically decrease.

3) MOMO is evaluated on the external dataset and compared to the Pan-Importer [16], a commercial product performing the classification task at our facility. An ablation study is performed, where the networks' decision power is increased by moving the network higher in the decision hierarchy. Additionally, we test MOMO without neural network support. For every algorithm, we report the percentage of correct predictions, minor and major errors, accuracy, and network contribution, on the external test set. An error is considered minor if the classification is correct but slightly too general (e.g. predicting "MR Spine" instead of "MR Thoracic Spine"), since this will not cause unnecessary examinations to be performed or studies not to be found. This error grading is performed automatically. All minor error combinations are listed in the Supplementary Materials.

## 2.7 Ethical Approval

This study was approved by the IRB of the University Hospital Essen (Application-Nr.: 20-9745-BO), explicit consent was waived due to its retrospective and technical nature.

# 3 Data Sharing

The code will be made publicly available upon full publication of this work.

# 4 Results

1) We evaluated multiple neural networks for different imaging modalities. The accuracies for these networks are shown in Table 1 and compared to the state-of-the-art (SOTA), if available. We achieved comparable performance across all modalities. Conventional radiographs achieved the highest prediction accuracy (97.1%, CI: 96.2%-98.0%), outperforming the current SOTA [7], and ultrasounds the lowest (81.4%, CI: 76.7%-86.1%).

2) Figure 3 shows the results of this simulation. The non-linear gain in accuracy through the voting mechanism is apparent, and quite large even for low network accuracies. We also find that the temperature scaling improves the performance of the voting concept. To understand why, observe the performance line for 2 series per study at a network per-image-series accuracy of 50% in Figure 3. Naively, one would expect 50% accuracy on the study level as well, as the network can predict either one or the other. However, as the network is calibrated, the predicted probability for a vote holds additional information. If the vote has a high probability, it is likely to be a good vote in practice, and a bad one for low probability. This increases our accuracy. We additionally see that a network which generalizes well to all test data, will yield a significant study prediction accuracy even for a comparatively low single-image prediction accuracy.

3) We also explored the study classification performance of MOMO, testing a wide range of settings and methods to combine metadata and image information. The results of the performance evaluation can be seen in Fig. 4. The network ensemble alone achieved 99.05% predictive power, 72.69% accuracy, and 10.3% minor errors. The commercial product reached 96.20% predictive power, 82.86% accuracy, and 1.36% minor errors.

Overall, we found that a hybrid version of MOMO yielded the best performance, utilizing metadata and neural networks in a single large vote. It outperformed other variants by a significant margin, scoring 99.29% predictive power, 92.71% accuracy, and 2.61% minor errors (using the networks with the best cross-validation performance).

| Modality | ResNet-152 | DenseNet-161 | SOTA in literature |
|----------|-----------|--------------|--------------------|
| CR | 96.5%<br>(CI: 95.0%-98.0%) | **97.1%**<br>**(CI: 96.2%-98.0%)\*** | 90.3%<br>(CI: 89.2%-91.3%) |
| CT | 84.6%<br>(CI: 81.7%-87.5%) | 87.7%<br>(CI: 85.2%-90.2%)\* | **91.9%**<br>**(CI: 90.2%-92.1%)** |
| MRI | 80.2%<br>(CI: 77.4%-82.9%) | 85.4%<br>(CI: 83.4%-87.5%)\* | **94.2%**<br>**(CI: 92.0%-95.6%)** |
| US | 79.0%<br>(CI: 75.0%-83.0%) | **81.4%**<br>**(CI: 76.7%-86.1%)\*** | - |
| XA | **83.5%**<br>**(CI: 77.3%-89.6%)\*** | 81.2%<br>(CI: 76.6%-85.9%) | - |

Table 1: Per-series classification accuracy of the neural networks. Accuracy and 95% confidence interval (CI) are reported. If applicable, a comparison with the best known comparable literature value is made (note that these have different underlying scopes, training, and test sets). All of our results are derived from five-fold cross-validation. Networks marked with a (\*) symbol are kept for MOMO.

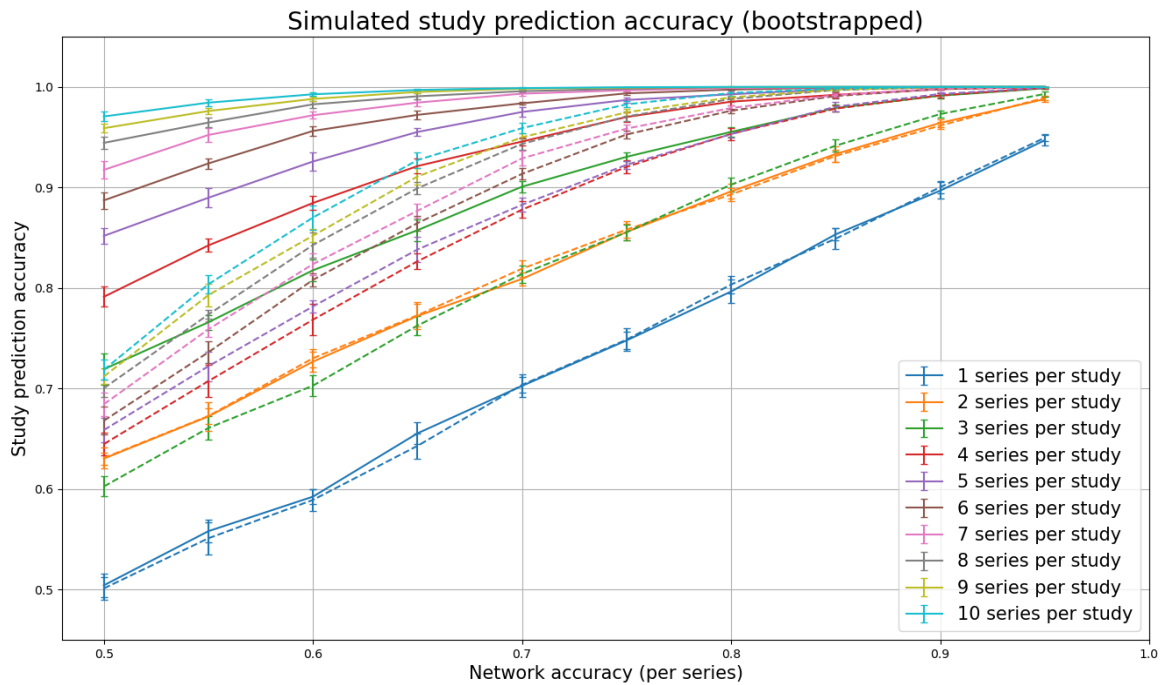

Figure 3: Plotted are classification results for simulated pseudostudies. The X-axis displays the accuracy of the simulated network's predictions, the Y-axis displays the resulting classification accuracy of the pseudostudies. The different lines are color-coded, showing varying numbers of series per study. A full line indicates that false predictions are uncorrelated. A dashed line indicates that false predictions are maximally correlated. The plot shows the mean accuracy across one million pseudostudies and the standard deviation.

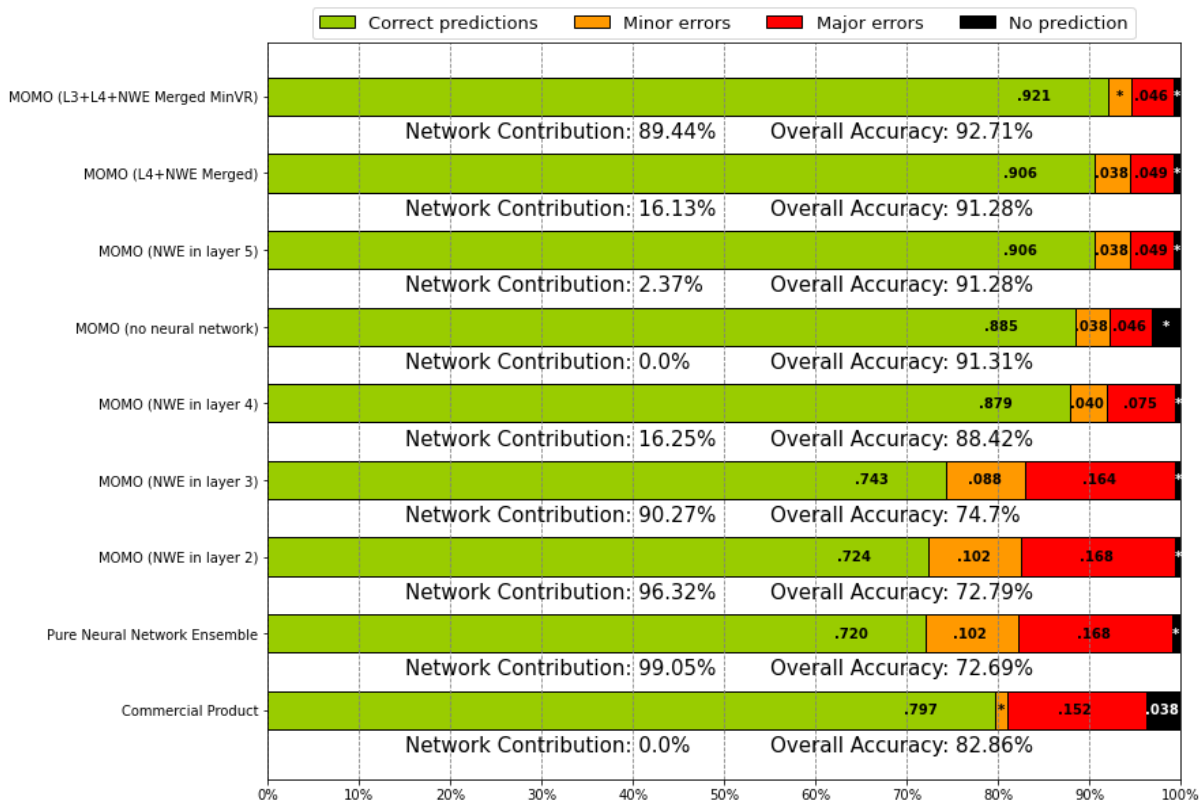

Figure 4: The figure displays correct predictions, minor and major errors for different study classifiers. Included are MOMO with and without network ensemble (NWE) with varying decision power (L5 denotes that the neural network is in the fifth (last) layer of MOMO, L1 would be the first layer), as well as a classifier based purely on neural networks and the commercial product. Additionally, two hybrid variants, where the network votes together with other layers (merged votes), are tested. On the "Minimal Vote Rules" (MinVR) setting, fewer rules for vote modification (see Chapter 2.4) were applied for technical reasons. We report classification accuracy, number of correct and incorrect predictions (split by severity of error), and the contribution of the networks.

# 5 Discussion

Imaging studies from external hospitals are often composed and named differently from the ones locally conducted. This generally necessitates manual identification of such studies. In this study, we presented MOMO, a metadata-driven algorithm, supported by an ensemble of deep neural networks. It was trained to recognize body regions and automatically classify such studies, and we justified its voting-based prediction mechanism in a separate Monte Carlo experiment. We found that the network ensemble alone can perform similarly to a commercial product, while MOMO outperformed either. We found that the neural networks offered a boost in predictive power and the best performance on our task was achieved with a variant of MOMO utilizing neural networks.

The state-of-the-art for comparable single series prediction is given by Dratsch et al. [7] for plain radiographs and Raffy et al. [8] for CT and MRI, while no deep learning body-region classification exists for Ultrasound or X-Ray Angiography. Our training results compare favorably to the SOTA for radiographs. For CT and MRI, our accuracies are comparatively lower but remain somewhat competitive. There are multiple explanations for this. The training set in [8] covered a larger domain (multiple hospitals, scanner types, etc.) and contained more images, likely allowing for better generalization. Beyond that, as our labels come from the PACS and not manual labeling, some labels may either be false (an abdomen study containing also a series of the chest, which was consequently labeled "abdomen") or correct but difficult to identify (such as a Vessel CT, which can have examples of various body sites in its corresponding studies). Since the latter problems are intrinsic to the compositionality of the study classification problem, and all three studies are somewhat different in scope, one could also argue that the comparison is not valid.

We observed that the network ensemble offered an improvement over no-network variants of MOMO (see Fig. 4), despite not performing as strongly on the external data as during training. However, this improvement only applied if the networks were not given too much decision power or if the network and metadata votes were combined for more robustness. Finally, we note that, qualitatively, the erroneous predictions made by the neural networks have an increased tendency to be anatomically related to the truth, compared to the metadata-driven approaches.

This work has limitations. If many similar keywords are added to the algorithm's reference database, it can cause mispredictions to increase because of false positives. Some classes are underrepresented during training, potentially worsening generalization. Additionally, the dataset used for training is automatically labeled, which may decrease the performance of the neural networks. Similarly, the compositionality of different types of studies poses problems for neural networks that learn to predict based on single series. Finally, the choice (single institution) and amount of training data is a limiting factor.

In the future, the use of additional training data (both local as well as external data) will allow the networks to mitigate some of these limitations. Additionally, new preprocessing steps (like contrast enhancement, cropping, etc.) could be introduced. Furthermore, improvements could be made by incorporating new decision layers into MOMO or improving the string matching (by comparing string similarity between metadata and keywords using metrics such as Levenshtein distance). Another interesting approach may be to perform predictions directly on the study level, using a Multiple Instance Learning classifier, which is particularly well-suited to solve the problem of compositionality we experienced classifying DICOM studies.

In Summary, this is the first open-source study classification tool to use metadata and neural networks for its decision process. It is fully automated and covers all common radiological modalities, leading to increased quality, and thus patient safety, as well as reduced workload for the clinical staff. Hence, the key results of this contribution are:

1) The algorithm can successfully identify 76 medical study types across seven modalities (CT, X-Ray Angiography, Radiographs, MRI, PET (+CT/MRI), Ultrasound, and Mammograms).

2) The algorithm outperforms a commercial product performing the same task by a significant margin.
3) The algorithm's performance increases through the application of deep learning techniques.

# Acknowledgements

This work received support from the Cancer Research Center Cologne Essen (CCCE).

# References

[1] Codari, M. et al. (2019). Impact of artificial intelligence on radiology: a EuroAIM survey among members of the European Society of Radiology. *Insights into imaging*, **10, 1-11**. doi: 10.1186/s13244-019-0798-3.

[2] Wang, J. X., Sullivan, D. K., Wells, A. J., Wells, A. C., & Chen, J. H. (2019). Neural networks for clinical order decision support. *AMIA Summits on Translational Science Proceedings*, **2019, 315**. url: https://pubmed.ncbi.nlm.nih.gov/31258984. pmid: 31258984.

[3] Byun, S. S. et al. (2021). Deep learning based prediction of prognosis in nonmetastatic clear cell renal cell carcinoma. *Scientific reports*, **11(1), 1-8**. doi: 10.1038/s41598-020-80262-9.

[4] Zaharchuk, G. (2020). Fellow in a Box: Combining AI and Domain Knowledge with Bayesian Networks for Differential Diagnosis in Neuroimaging. *Radiology*, **295(3), 638-639**, doi: 10.1148/radiol.2020200819.

[5] Bidgood Jr, W. D., Horii, S. C., Prior, F. W., & Van Syckle, D. E. (1997). Understanding and using DICOM, the data interchange standard for biomedical imaging. *Journal of the American Medical Informatics Association*, **4(3), 199-212**. doi: 10.1136/jamia.1997.0040199. url: https://pubmed.ncbi.nlm.nih.gov/9147339. 9147339[pmid].

[6] Gueld, M. O. et al. (2002, May). Quality of DICOM header information for image categorization. In *Medical imaging 2002: PACS and integrated medical information systems: design and evaluation* **(Vol. 4685, pp. 280-287)**. International Society for Optics and Photonics. doi: 10.1117/12.467017.

[7] Dratsch, T. et al. (2021). Practical applications of deep learning: classifying the most common categories of plain radiographs in a PACS using a neural network. *European Radiology*, **31(4), 1812-1818**. doi: 10.1007/s00330-020-07241-6.

[8] Raffy, P. et al. (2021). Deep Learning Body Region Classification of MRI and CT examinations. Preprint at arXiv:2104.13826. (Manuscript submitted for publication).

[9] Zhang, P., Wang, F., & Zheng, Y. (2017, April). Self supervised deep representation learning for fine-grained body part recognition. In *2017 IEEE 14th International Symposium on Biomedical Imaging (ISBI 2017)* **(pp. 578-582)**. IEEE. doi: 10.1109/ISBI.2017.7950587.

[10] Sugimori, H. (2018). Classification of computed tomography images in different slice positions using deep learning. *Journal of healthcare engineering*, **Vol. 2018**. doi: 10.1155/2018/1753480.

[11] Yan, K., Lu, L., & Summers, R. M. (2018, April). Unsupervised body part regression via spatially self-ordering convolutional neural networks. In *2018 IEEE 15th International*

*Symposium on Biomedical Imaging (ISBI 2018)* **(pp. 1022-1025)**. IEEE.   doi: 10.1109/ISBI.2018.8363745

[12] K. He, X. Zhang, S. Ren and J. Sun, "Deep Residual Learning for Image Recognition," *2016 IEEE Conference on Computer Vision and Pattern Recognition (CVPR)*, **2016, pp. 770-778**, doi: 10.1109/CVPR.2016.90.

[13] G. Huang, Z. Liu, L. Van Der Maaten and K. Q. Weinberger, "Densely Connected Convolutional Networks," *2017 IEEE Conference on Computer Vision and Pattern Recognition (CVPR)*, **2017, pp. 2261-2269**, doi: 10.1109/CVPR.2017.243.

[14] Deng, J. et al. (2009, June). Imagenet: A large-scale hierarchical image database. In *2009 IEEE conference on computer vision and pattern recognition* **(Vol. 2009, pp. 248-255)**. IEEE. doi: 10.1109/CVPR.2009.5206848.

[15] Guo, C., Pleiss, G., Sun, Y., & Weinberger, K. Q. (2017, July). On calibration of modern neural networks. In *International Conference on Machine Learning* **(Vol. 2017 pp. 1321-1330)**. PMLR. doi: 10.5555/3305381.3305518

[16] Pansoma GmbH, url: https://www.pansoma.at/produkte/pan-proworx/pan-import/, last accessed 9th of August, 2021.

[17] Ioffe, S., & Szegedy, C. (2015, June). Batch normalization: Accelerating deep network training by reducing internal covariate shift. In *International conference on machine learning* **(Vol. 2015, pp. 448-456)**. PMLR. doi: 10.5555/3045118.3045167

# List of Abbreviations

**CR** - Conventional Radiograph
**CT** - Computed Tomography
**DICOM** - Digital Imaging and Communications in Medicine
**MG** - Mammogram
**MOMO** - Modality Mapping and Orchestration (the algorithm designed in this paper)
**MRI** - Magnetic Resonance Imaging
**PACS** - Picture Archiving and Communication Systems (most medical images are archived on such systems)
**PET** - Positron Emission Tomography
**US** - Ultrasound
**XA/DSA** - X-Ray/Digital Subtraction Angiography

# Supplementary Materials

## Modification or disallowing of votes

The votes of some DICOM series can be disallowed or modified according to a set of pre-specified rules found in the configuration file. Votes can be disallowed if the modality of the series does not match that of the study, or if blacklisted terms like "screenshot" exist in the *Series Description*, which indicate that the series may be non-representative of the whole study. Votes can be modified if some compositions of series are detected. For example, if a study contains both "CT Abdomen" and "CT Thorax" labels, all votes for "CT Abdomen" and "CT Thorax" are replaced by votes for "CT Thorax+Abdomen". This ruleset can be easily modified in the configuration files to represent different class compositions. As a rule of thumb, there exists a sweet spot of classification accuracy for a set of rules that is neither too large, nor too small.

## Training Hyperparameter Choice

The neural networks were all trained using the same set of hyperparameters. These were found using a non-exhaustive manual search and may not represent the optimal solution. However, we found these parameters to converge reliably no matter the network architecture or training set, with the accuracy not changing significantly for similar parameter settings and becoming worse (or not converging) for choices different by at least an order of magnitude.

All training was performed with batches of size B = 24, a learning rate of $\lambda_{nominal} = 1x10^{-4}$ and an exponential learning rate decay policy where $\lambda_{network} = \lambda_{nominal} * 0.98^{epoch}$.

All last batches are dropped (as not dropping differently sized batches results in worse performance when using BatchNorm [17]), and all gradients are left entirely unclipped. A weight decay of $\omega_{decay} = 1x10^{-4}$ is additionally used as a regularizer for the weights to help prevent overfitting on less well-represented features or classes in training. We perform no oversampling of underrepresented classes during training.

# Supplementary Table 1a: Examined Study Classes

| Study Type | Modality | # of instances in training |
|------------|----------|----------------------------|
| CT Abdomen | CT | 193 |
| CT Heart | CT | 171 |
| CT Lower Extremities | CT | 182 |
| CT of the Vessels | CT | 189 |
| CT Pelvis | CT | 190 |
| CT Skull | CT | 82 |
| CT Skull + Neck | CT | 150 |
| CT Spine | CT | 170 |
| CT Thorax | CT | 191 |
| CT Thorax + Abdomen | CT | 140 |
| CT Upper Extremities | CT | 197 |
| CT Whole Body | CT | 189 |
| DSA Abdomen | XA | 48 |
| DSA Angiology | XA | 15 |
| DSA Brain Vessels | XA | 0* |
| DSA Extremities | XA | 12 |
| DSA Liver | XA | 51 |
| DSA Pelvis/Upper Legs | XA | 93 |
| DSA Skull | XA | 36 |

## Supplementary Table 1b: Examined Study Classes (cont.)

| Study Type | Modality | # of instances in training |
|---|---|---|
| DSA Skull/Neck-Intervention | XA | 16 |
| DSA Spine | XA | 10 |
| DSA Thorax | XA | 18 |
| Myelography | XA | 174 |
| Screening of the Abdomen | XA | 37 |
| Screening of the Extremities | XA | 90 |
| Screening of the Spine | XA | 153 |
| Screening of the Thorax | XA | 61 |
| Conventional, Abdomen | CR | 186 |
| Conv., Arm | CR | 738 |
| Conv., Cervical Spine | CR | 177 |
| Conv., Foot | CR | 574 |
| Conv., Hand | CR | 382 |
| Conv., Legs | CR | 635 |
| Conv., Lumbar Spine | CR | 186 |
| Conv., Mammae | CR | 124** |
| Conv., Pelvis | CR | 189 |
| Conv., Skull | CR | 196 |
| Conv., Spine | CR | 0* |

## Supplementary Table 1c: Examined Study Classes (cont.)

| Study Type | Modality | # of instances in training |
|---|---|---|
| Conv., Thoracic Spine | CR | 188 |
| Conv., Thorax | CR | 185 |
| Mammography | MG | 124** |
| MRI Abdomen | MRI | 136 |
| MRI Cervical Spine | MRI | 157 |
| MRI Lower Extremities | MRI | 156 |
| MRI Lumbar Spine | MRI | 165 |
| MRI Mammae | MRI | 200 |
| MRI Pelvis | MRI | 194 |
| MRI Skull | MRI | 194 |
| MRI Skull + Neck | MRI | 198 |
| MRI Spine | MRI | 105 |
| MRI Thoracic Spine | MRI | 184 |
| MRI Thorax | MRI | 187 |
| MRI Upper Extremities | MRI | 144 |
| MRI Whole Body | MRI | 90 |
| PET CT Heart | PET + CT | 179*** |
| PET CT Lower Extremities | PET + CT | 172*** |
| PET CT Skull | PET + CT | 190*** |

# Supplementary Table 1d: Examined Study Classes (cont.)

| Study Type | Modality | # of instances in training |
|---|---|---|
| PET CT Skull + Neck | PET + CT | 166*** |
| PET CT Upper Extremities | PET + CT | 55*** |
| PET CT Whole Body | PET + CT | 189*** |
| PET MRI Abdomen | PET + MRI | 178*** |
| PET MRI Lower Extremities | PET + MRI | 160*** |
| PET MRI Mammae | PET + MRI | 181*** |
| PET MRI Skull | PET + MRI | 192*** |
| PET MRI Skull+Neck | PET + MRI | 198*** |
| PET MRI Spine | PET + MRI | 179*** |
| PET MRI Thorax | PET + MRI | 198*** |
| PET MRI Upper Extremities | PET + MRI | 186*** |
| PET MRI Whole Body | PET + MRI | 187*** |
| Ultrasound Abdomen | US | 199 |
| Ultrasound Extremities | US | 112 |
| Ultrasound Mammae | US | 83 |
| Ultrasound Neck | US | 111 |
| Ultrasound Skull | US | 5 |
| Ultrasound Thorax | US | 41 |
| Ultrasound Whole Body | US | 45 |

* Note that some classes are not represented in the neural network training. These classes are only predicted by the metadata-driven part of the algorithm.

** Mammographies themselves were not used to train a neural network (as there is only one single study type for the modality). They were used to train the CR-network, by labeling them as conventional radiographs of the mammae (which is what they are, ostensibly). No study not marked as modality MG (Mammography) and containing an image of the mammae

exists in any of our datasets (the modality of a study being a very reliable piece of metadata), so the point is purely technical.

*** Note that the modalities which contain PET images do not have a dedicated network. In the presence of a PET image, the MRI or CT network predicts the class as if there were no PET images and then picks the corresponding PET class as answer. Thus, the non-PET images of PET+MRI or PET+CT simply become part of the MRI or CT training sets.

# Supplementary Table 2a: List of minor error combinations

| True study class | Misclassification as X treated as minor |
|---|---|
| CT Abdomen | CT Thorax+Abdomen |
| CT Thorax | CT Thorax+Abdomen |
| CT Skull | CT Skull+Neck |
| DSA Abdomen | Screening of the Abdomen |
| DSA Brain Vessels | DSA Skull, DSA Skull/Neck-Intervention |
| DSA Extremities | Screening of the Extremities |
| DSA Liver | DSA Abdomen |
| DSA Spine | Screening of the Spine |
| DSA Skull | DSA Brain Vessels, DSA Skull-Neck Intervention |
| DSA Skull/Neck-Intervention | DSA Brain Vessels, DSA Skull |
| DSA Thorax | Screening of the Thorax |
| Screening of the Abdomen | DSA Abdomen |
| Screening of the Extremities | DSA Extremities |
| Screening of the Spine | DSA Spine |
| Screening of the Thorax | DSA Thorax |
| Conventional, Cervical Spine | Conv., Spine |
| Conv., Lumbar Spine | Conv., Spine |
| Conv., Thoracic Spine | Conv., Spine |

# Supplementary Table 2b: List of minor error combinations (cont.)

| True study class | Misclassification as X treated as minor |
|---|---|
| Conv., Mammae | Mammogram |
| PET CT Skull | PET CT Skull+Neck |
| PET MRI Skull | PET MRI Skull+Neck |
| MRI Cervical Spine | MRI Spine |
| MRI Lumbar Spine | MRI Spine |
| MRI Thoracic Spine | MRI Spine |
| MRI Skull | MRI Skull+Neck |

The thought process behind these minor errors is that it needs to cause no risk of additional but unnecessary examinations, to be classified as minor. If a doctor opened a patient's files and saw "MRI Spine", they would check if their requested lumbar spine MRI images are in that study. They might not even bother to check if it was labeled "MRI Skull" instead.